%% file: main.tex
\documentclass[10pt,twocolumn,letterpaper]{article}

\makeatletter
\@namedef{ver@everyshi.sty}{}
\makeatother

\newif\ifarxiv
\arxivtrue 

\ifarxiv
  \usepackage[pagenumbers]{cvpr}
  \newcommand{\ARXIVversion}[2]{#1}
\else
  \usepackage{cvpr}
  \newcommand{\ARXIVversion}[2]{#2}
\fi

\usepackage{graphicx}
\usepackage{tikz}
\usepackage{amsmath}
\usepackage{amssymb}
\usepackage{booktabs}
\usepackage{soul}
\usepackage{flushend}
\usepackage[accsupp]{axessibility}

\newcommand{\paragraphcustom}[1]{\vspace{1pt}\noindent\textbf{#1}}
\newcommand{\paragraphcustomWOvspace}[1]{\noindent\textbf{#1}}

\definecolor{cvprblue}{rgb}{0.21,0.49,0.74}
\usepackage[pagebackref,breaklinks,colorlinks,citecolor=cvprblue]{hyperref}

\usepackage[capitalize]{cleveref}
\crefname{section}{Sec.}{Secs.}
\Crefname{section}{Section}{Sections}
\Crefname{table}{Table}{Tables}
\crefname{table}{Tab.}{Tabs.}

\def\cvprPaperID{1309} 
\def\confName{CVPR}
\def\confYear{2024}

\begin{document}

\title{GenHowTo: Learning to Generate Actions and State Transformations \\ from Instructional Videos}

\author{
	Tom\'{a}\v{s} Sou\v{c}ek\textsuperscript{1}
	\quad\quad
	Dima Damen\textsuperscript{2}
	\quad\quad
	Michael Wray\textsuperscript{2}
	\quad\quad
	Ivan Laptev\textsuperscript{3}
	\quad\quad
	Josef Sivic\textsuperscript{1}
	\\
	\small{$^1$CIIRC CTU \quad $^2$University of Bristol \quad $^3$MBZUAI}
	\\
	\small{\texttt{tomas.soucek@cvut.cz}}
	\quad\quad
	\small{\url{https://soczech.github.io/genhowto/}}
}

\twocolumn[{
\maketitle

\vspace*{-1.2em}
\vspace*{-0.5em}
\centering
\includegraphics[width=0.96\textwidth]{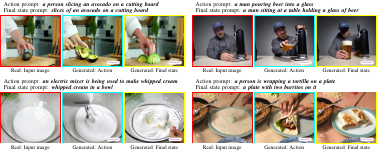}
\vspace*{-0.3cm}
\captionof{figure}{Given an image of an initial scene (\textcolor{red}{red}) and text prompts ({\bfseries{\textit{bold}}}), GenHowTo generates images corresponding to the action (\textcolor{cyan}{blue}) and the final state when action is completed (\textcolor{yellow}{yellow}). GenHowTo is learned from instructional videos and can generate new images of both seen and previously unseen object transformations. Importantly, GenHowTo learns to maintain the parts of the scene that showcase the action carried out in the same environment, in the spirit of HowTo examples, while introducing important objects (\eg, hand and knife in the first example) and transforming the object according to the prompt.
\vspace{.6cm}
}
\label{fig:teaser}
}]

\begin{abstract}
\mbox{}\vspace{-.8cm}\\
We address the task of generating temporally consistent and physically plausible images of actions and object state transformations. Given an input image and a text prompt describing the targeted transformation, our generated images preserve the environment and transform objects in the initial image. Our contributions are threefold. First, we leverage a large body of instructional videos and automatically mine a dataset of triplets of consecutive frames corresponding to initial object states, actions, and resulting object transformations. Second, equipped with this data, we develop and train a conditioned diffusion model dubbed GenHowTo. Third, we evaluate GenHowTo on a variety of objects and actions and show superior performance compared to existing methods. In particular, we introduce a quantitative evaluation where GenHowTo achieves 88\% and 74\% on seen and unseen interaction categories, respectively, outperforming prior work by a large margin.
\end{abstract}



\section{Introduction}
\label{sec:intro}

\footnotetext[1]{Czech Institute of Informatics, Robotics and Cybernetics at the Czech Technical University in Prague.}
\footnotetext[3]{Mohamed bin Zayed University of Artificial Intelligence.} 
Imagine you want to cook dinner or prepare your favorite cocktail to spend an evening with friends.
Your inspiration may come from an image of a dish you have recently seen in a restaurant or a cookbook. 
In a similar spirit, recent work in robotics aims to learn goal-conditioned policies where the tasks or intermediate goals are defined by images~\cite{du2023learning,pathakICLR18zeroshot,nair2018visual,nair2020contextual,nasiriany2019planning,wang2023manipulate}. Image-defined tasks have an advantage of providing detailed information such as
the desired stiffness of the whipped cream or the thickness of avocado slices, which can be difficult to convey by language alone. 

Moreover, policies can be further improved by image sequences illustrating the execution of a task in terms of actions~\cite{chane2023learning,chen2021learning}.
However, images of target states and actions are rarely available in advance for arbitrary tasks and environments.
Motivated by this limitation, in this work, we aim to generate realistic images of actions and target states for a variety of tasks defined by text prompts, while {\em preserving the same environment} across the initial image and the generated images of the future, see Figure~\ref{fig:teaser}.

With the advent of powerful vision-language models, recent work excels in generating realistic and high-fidelity images from textual descriptions~\cite{liu2022compositional,gafni2022make,saharia2022photorealistic,gal2022image,ruiz2023dreambooth,ramesh2022hierarchical}. 
Other recent methods show impressive results for text-conditioned image editing~\cite{hertz2022prompt,mokady2022null,huberman2023edit,brooks2023instructpix2pix,kawar2023imagic} where input images are modified according to provided text prompts.
As these models are not trained to focus on action-induced transformations, they may fail to maintain a consistent environment while generating visually small yet semantically complex object changes.
To address this challenge, we leverage image sequences from large-scale video data for training generative models. In particular, we exploit instructional videos with variation of tasks, scenes, and object state changes.

Instructional videos, however, lack annotation at scale. 
To this end, we build on the recent work~\cite{soucek2022multi} and automatically create a large-scale dataset with 200k image triplets and corresponding textual descriptions. Our triplets are temporally ordered video frames corresponding to (i)~initial object states, (ii)~state-modifying actions, and (iii)~objects in new states. Equipped with this dataset, we propose a text- and image-conditioned generative model, GenHowTo, that transforms input images into images with actions and new object states according to textual prompts. 
We perform qualitative and quantitative evaluation of our method and demonstrate its superior performance.

In summary, we make the following contributions. 
(1)~We address the problem of generating physically plausible state transformations of input images according to a variety of action and final state prompts and automatically collect a large-scale training dataset from narrated instructional videos.
(2)~We propose GenHowTo, a text-conditioned image generation model that modifies input images according to action or final state text prompts while preserving the scene from the input image. 
(3)~We evaluate GenHowTo both qualitatively and quantitatively and show its superior performance compared to the state of the art.  
For quantitative results, we train classifiers on object states and actions to assess the method's ability to transform an image from its initial state to the action or final states.
Our code, data, and pre-trained models are publicly available.

\begin{figure*}[t]
\centering
\includegraphics[width=1.0\linewidth]{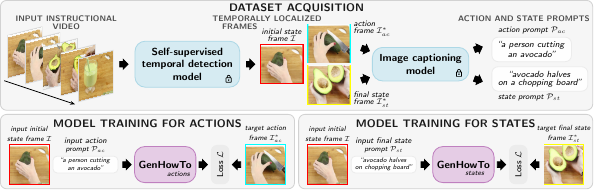}
  \vspace*{-6mm}
\caption{\textbf{Method overview.} We use a self-supervised model to detect objects before, during, and after they are manipulated in instructional videos (top left). Then, the detected frames are automatically annotated using an image captioning model (top right). Finally, the detected frames with the text annotations are used to train our two diffusion models for transforming objects in the images (bottom).}
\label{fig:overview}
\vspace*{-2mm}
\end{figure*}

\section{Related work}
\label{sec:related_work}

\paragraphcustomWOvspace{Object states.} Localization, recognition, and generation of object transformations, and changes of object states, play a key role in applications such as procedural planning, robotics, and video action understanding~\cite{du2023learning,zhong2023learning,zhao2022p3iv,wang2023manipulate}.
Most works focus on recognition, compositionality, and localization in videos~\cite{soucek2022look,doughty2020action,naeem2021learning,epstein2021learning,dvornik2023stepformer}; however, the generation of object states has been underexplored. While recent powerful diffusion models have been able to generate rich, diverse, and high-quality images, generating objects in exact states (\eg, carrot cut in a baton shape) is still a challenge \cite{saini2023chop}. Saini \etal \cite{saini2023chop} introduces a novel benchmark for the generation of object states, yet their focus is very limited to only the \textit{cutting} action and a small dataset of objects recorded against a green screen. We focus on transforming a much wider range of actions and objects in realistic scenes by using in-the-wild instructional internet videos as the source of training data.

\paragraphcustom{Conditional generation.}
GANs~\cite{xu2018attngan,mirza2014conditional,odena2017conditional} and auto-regressive models~\cite{ramesh2021zero,esser2021taming} allowed conditional synthesis of images.
Recently, large-scale diffusion models have enabled the synthesis of high-resolution realistic images with diverse and complex attributes~\cite{bansal2023universal,yang2023paint,dhariwal2021diffusion,ho2020denoising,song2020denoising,liu2022compositional,rombach2022high,saharia2022palette,saharia2022photorealistic,wang2022pretraining,gafni2022make,gal2022image,ruiz2023dreambooth, ramesh2022hierarchical,kingma2021variational}. 
These models may be conditioned on images~\cite{saharia2022palette,wang2022pretraining,yang2023paint}, textual descriptions~\cite{liu2022compositional,gafni2022make,saharia2022photorealistic,gal2022image,ruiz2023dreambooth,ramesh2022hierarchical}, or other modalities~\cite{bansal2023universal} to produce the desired output.
While such models produce images that satisfy the text prompt or are similar to the input image, applying any targeted edits is difficult. To tackle that, new text-based methods and image-to-image models for image editing have emerged.

\paragraphcustom{Text-based image editing.} Using visual-text embeddings, such as CLIP~\cite{radford2021learning}, allowed editing images with only textual descriptions~\cite{tumanyan2023plug,kim2022diffusionclip,nichol2021glide,avrahami2022blended,parmar2023zero,patashnik2023localizing,hertz2022prompt,mokady2022null,kawar2023imagic,huberman2023edit}.
Prompt-mixing~\cite{patashnik2023localizing} uses different text prompts during the de-noising process to generate image variations.
Alternatively, Prompt-to-Prompt~\cite{hertz2022prompt} can apply various edits to generated images by tweaking activations of a trained diffusion model. Further, it can also be extended to editing real images~\cite{mokady2022null}.
Similarly, Imagic~\cite{kawar2023imagic} also edits images using text prompts.
However, these methods require hundreds of back-propagation steps to edit a single image, making them impractical.
In contrast, we train a diffusion model that can manipulate any image by applying a standard diffusion schedule with the source image and the target prompt.

\paragraphcustom{Image-to-image models.}
Image-to-Image translation~\cite{isola2017image,zhu2017unpaired,zhang2023adding,brooks2023instructpix2pix,choi2018stargan,park2019semantic,wang2018high,zhang2020cross,zhou2021cocosnet,zhu2017toward,chen2021pre,esser2021taming,ramesh2021zero} attempts to convert images between domains.
Image-conditioned image generation has been attempted with adversarial networks using paired~\cite{isola2017image} or unpaired data with cycle-consistent loss~\cite{zhu2017unpaired}. Other methods explored conditional generation by inpainting a masked-out part of the input image~\cite{tao2013error,chen2019toward,guo2021intrinsic,yang2023paint, kulal2023putting}.
Recently, Zhang~\etal introduced ControlNet~\cite{zhang2023adding}---a scheme for training diffusion models with additional image conditioning using paired data such as automatically obtained edge images, thus allowing for fine-grained image generation.
However, adapting diffusion models to image-to-image generation with unpaired data is not trivial.
Brooks~\etal \cite{brooks2023instructpix2pix} solves the lack of paired data by generating the pairs using diffusion models.
In contrast, we find corresponding pairs of images by mining internet instructional videos.

\paragraphcustom{Video Frame Prediction.}
Another related topic is video frame prediction, in which future frames of a video are generated.
This task has been addressed using recurrent networks~\cite{srivastava2015unsupervised,shi2015convolutional,lotter2017deep}, adversarial training~\cite{mathieu2015deep,luo2021future,liang2017dual,kwon2019predicting}, or, more recently, diffusion  models~\cite{yan2023feature}.
We differ from these approaches as we predict semantic change, such as the action and the final state, using a text prompt, not a continuous frame sequence immediately following the input frames.

\section{Learning to generate actions and state transformations}
\label{sec:method}

Given an image of an object in a scene, our goal is to generate images that showcase an \textit{action exerted} upon the object as well as the \textit{transformation of the object} as a result of the outcome of that action. 
Importantly, we wish to produce consistent images, where the action is exerted on the same object in an unchanged environment.
For this conditional generation, three challenges need to be tackled:

First, from the same input image, many possible actions or object transformations can be generated. It is essential to condition the generation process on this sought-after transformation.
We address this challenge by specifying the action and the desired final state of the object through given text prompts.
For example, given an image of an avocado on a chopping board, we specify the action through the prompt: \textit{``a person cutting an avocado''} and the desired final state via the prompt: \textit{``avocado halves on a chopping board.''}

Second, to generate consistently paired images showcasing the action or object transformation, the generation should maintain parts of the scene unrelated to the action, including the background and other objects (\eg, the chopping board in the example above) but also introduce new objects consistent with the action (\eg, a knife and a person's hand). 
We address this challenge with an image-conditioned diffusion model fine-tuned for maintaining a consistent background between pairs of images. 

Third, to train an image-conditioned model, data of such paired images are required. Manually collecting such data at scale is very challenging due to the nature of the problem.
We address this challenge by mining 200k triplets of images from instructional HowTo videos.

We define our problem formally in Section~\ref{subsec:formal_def}. Then, we address the challenge of acquiring paired image data by extracting images from instructional videos in Section~\ref{subsec:dataset_acq}. In the same section, we address the issue of many possible object transformations by automatically collecting text prompts describing the transformations. Lastly, in Section~\ref{subsec:genhowto}, we describe our image- and text-conditioned model to modify the object but maintain the scene.

\subsection{Problem definition}
\label{subsec:formal_def}
We define the problem as follows. Let $\mathcal{I}$ be an image depicting the object in its initial state and $\mathcal{P}$ be a text prompt describing an action exerted upon an object or a transformation (the final state) of the object in the image~$\mathcal{I}$. Then, we define the target image $\mathcal{I}^*$ as the modification of $\mathcal{I}$ that satisfies the prompt~$\mathcal{P}$. Our goal is to produce an image similar to $\mathcal{I}^*$ given the original image~$\mathcal{I}$ and the text prompt~$\mathcal{P}$. We differentiate between two target images~$\mathcal{I}^*$---one that shows the ongoing action, denoted as~$\mathcal{I}^*_{ac}$, and another that demonstrates the final state 
after the action has concluded, denoted as~$\mathcal{I}^*_{st}$. Similarly, we differentiate the prompts describing the action $\mathcal{P}_{ac}$ and the final state $\mathcal{P}_{st}$. We omit the subscripts if we talk about both actions and the final states. We use the input image $\mathcal{I}$ and one of the prompts $\mathcal{P}$ as input. The corresponding target image $\mathcal{I}^*$ to the text prompt serves as the ground truth in the loss during training.

\subsection{Dataset acquisition}
\label{subsec:dataset_acq}
To form the dataset, our goal is to acquire 5-tuples $(\mathcal{I}, \mathcal{I}^*_{ac}, \mathcal{P}_{ac}, \mathcal{I}^*_{st}, \mathcal{P}_{st})$ which include the input image, the target action image, the action text prompt, the target final-state image, and the final-state text prompt, respectively.
For example, the input image can be an empty glass on a countertop. For the action of filling the glass with orange juice, we aim to find two images, one showcasing the action of filling the glass and a second image showcasing the end state after the action is finished. Importantly the two images should showcase the same glass on the same countertop.

While it is possible to obtain images of glasses, both empty and full, on kitchen countertops, obtaining triplets of images showing (i) the initial state, (ii) the action, and (iii) the final state with the same glass and background in sufficient quantity is difficult. 
This correspondence, while missing from arbitrary images, is present in instructional videos. Such videos often capture visual changes of objects in static and continuous shots, \ie, keep the same background.
However, the challenge in using instructional videos is to find the right frames that correspond to such triplets.
We address this challenge next.

\paragraphcustom{Image triplets from instructional videos.}
We seek instructional videos as a source for our corresponding image triplets.
Instructional videos have the advantage of typically capturing a procedure with minimal camera and scene changes, thus showcasing object changes while preserving the background.
To get image triplets from these videos, we use the self-supervised model of Sou\v{c}ek \etal~\cite{soucek2022multi}, which localizes actions and objects in various states from videos automatically. The model is trained with sets of videos from various interaction categories such as \textit{chopping onions} or \textit{assembling a PC}. For each video frame, the model predicts the likelihood of the frame containing the object (\eg, onion or PC) in its initial state (before the action), during the action, or in its final state (after the action). 
The model is trained in a self-supervised iterative process where the target labels are computed from the model's own predictions subject to ordering and other constraints. We use this method to train the model on the COIN~\cite{tang2019coin} and the ChangeIt~\cite{soucek2022look} datasets.
Once trained, we use this model to localize the image triplets in the input videos.

In detail, for every video from the two datasets, we apply the trained model to select the frames with the highest initial state, action, and final state scores.
From each video, we use these three frames to form the triplet. In the next paragraphs, these are extended with text prompts $\mathcal{P}_{ac}$ and  $\mathcal{P}_{st}$ to form the dataset 5-tuple.

\paragraphcustom{Enriching images with text prompts.}
For each input image $\mathcal{I}$, there are multiple possible outcomes. For example, if the input image contains an avocado, the outcome may be the avocado in halves or slices, peeled or smashed.
As we wish to generate the image of the action or object transformation, we need a way to specify which of the various options the model should target.
Building on the success of text-based conditioning, we use a text description of the target image to guide the image generation.
We refer to this text description as the text prompt $\mathcal{P}$.

A straightforward method of obtaining the text prompt could utilize the corresponding narration from the instructional video where the images were obtained. However, it has been shown that automatic narrations do not always correspond to the ongoing action~\cite{Han_2022_CVPR,Miech_2019_ICCV}. Additionally, even when aligned, we observe the narrations do not emphasize the objects and actions shown in videos (see examples in the \ARXIVversion{appendix}{appendix~\cite{soucek24genhowto}}). Therefore, instead of the narrations, we use the image captioning model BLIP2~\cite{li2022blip} applied to both target images $\mathcal{I}^*_{ac}$ and $\mathcal{I}^*_{st}$ to produce accurate text prompts $\mathcal{P}_{ac}, \mathcal{P}_{st}$ describing the desired outcome of the transformation, completing the dataset 5-tuple~$(\mathcal{I}, \mathcal{I}^*_{ac}, \mathcal{P}_{ac}, \mathcal{I}^*_{st}, \mathcal{P}_{st})$.

\subsection{GenHowTo model}
\label{subsec:genhowto}
We propose a model for generating our target images using the input image and text prompts.
We refer to our model as~\textbf{GenHowTo}, given its capability of generating actions or final state images from an initial image and the text prompt, as well as its training from instructional videos.

To train our GenHowTo model, we utilize the dataset described in Sec.~\ref{subsec:dataset_acq}.
Critically, we train two distinct models with different sets of weights, one for generating action images and one for generating the final state images. We ablate this decision against training jointly in Section~\ref{sec:experiments}.
We thus instantiate two subsets from our dataset: $(\mathcal{I}, \mathcal{I}^*_{ac}, \mathcal{P}_{ac})$ for actions and $(\mathcal{I}, \mathcal{I}^*_{st}, \mathcal{P}_{st})$ for final states.
For the rest of this section, we refer to the input to our model in a general manner as $(\mathcal{I}, \mathcal{I}^*, \mathcal{P})$.
The notation is applicable to any of the two cases noted above.

Our model aims to generate images that maintain parts of the input scene irrelevant to the action or object transformation, such as the background. At the same time, the model needs to modify objects in the scene while preserving their identity or even introduce new objects in accordance with the text prompt. For example, the new objects can be hands, if a prompt calls for some action performed. To solve these challenges, we build on the state-of-the-art ControlNet architecture \cite{zhang2023adding}, which we adapt to our setting. 
In particular, in order to preserve the input scene yet allow for major semantic edits (introducing new objects or changes of the object state), we (i) introduce semantic conditioning on the input image at training time and (ii) initialize the noise sampling from the input image at inference time. Details of the model, including our modifications, are given in the remainder of this section. 

\begin{figure}[t]
\centering
\includegraphics[width=1.0\linewidth]{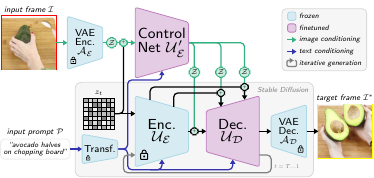}
\vspace{-0.6cm}
\caption{\textbf{GenHowTo model overview.} {
The model $\epsilon_\theta$ takes as input (left) a frame depicting the object in its initial state $\mathcal{I}$ and a text prompt $\mathcal{P}$ describing an action or the desired final state. The output of the model is an image $\mathcal{I}^*$ of the same scene but depicting the action or the desired final state. 
}}
\label{fig:model}
\vspace{-0.2cm}
\end{figure}

\paragraphcustom{Model components.}
The model $\epsilon_\theta$, shown in Figure~\ref{fig:model}, is a pretrained text-conditioned latent Stable Diffusion~\cite{rombach2022high} formed by latent U-Net encoder $\mathcal{U}_\mathcal{E}$ and latent U-Net decoder $\mathcal{U}_\mathcal{D}$. The latent representation is then decoded into the output image using a VAE decoder $\mathcal{A}_\mathcal{D}$. To incorporate the conditioning on the input RGB image, the latent U-Net encoder $\mathcal{U}_\mathcal{E}$ is duplicated to form the ControlNet encoder $\mathcal{U}'_\mathcal{E}$. The outputs of both encoders are summed together before being passed into the UNet decoder $\mathcal{U}_\mathcal{D}$.

\paragraphcustom{Semantic image conditioning.}
While ControlNet has mainly focused on spatial edits by conditioning, \eg, on edge maps or human poses, our goal is to enable semantic edits 
of the input scene given by an RGB image. Hence, in contrast to ControlNet \cite{zhang2023adding}, which projects the input image into the latent space by a handful of randomly initialized layers, our model directly projects the input image into the latent ``semantic" space of the U-Net using the encoder~$\mathcal{A}_\mathcal{E}$ of the pre-trained VAE.
This semantic representation of the input image is then summed with the standard noise input and passed into the ControlNet encoder. 
To produce meaningful gradients at the start of the training, independent learnable convolution layers $\mathcal{Z}_i$ are added to the outputs of the ControlNet encoder $\mathcal{U}'_\mathcal{E}$ with their parameters initialized to zeros.

\paragraphcustom{Model training.}
All parameters, including the duplicated ControlNet encoder $\mathcal{U}'_\mathcal{E}$, are initialized by the pretrained Stable Diffusion weights except for the zeroed-out convolutional layers $\mathcal{Z}_i$. During training, we keep the original Stable Diffusion encoder $\mathcal{U}_\mathcal{E}$ frozen and only finetune the decoder $\mathcal{U}_\mathcal{D}$ and the ControlNet encoder $\mathcal{U}'_\mathcal{E}$.
For training, the target image $\mathcal{I}^*$ is encoded by the pretrained VAE encoder to produce the latent encoding $z$. 
Further, a timestep $t\in [1, T]$ is sampled, and noise $\epsilon$ is added to $z$ according to the diffusion schedule, producing the noised latent $z_t$.
The model $\epsilon_\theta$ is trained to minimize the squared $L_2$ loss (Eq.~\eqref{eq:loss}) between the predicted noise and the actual noise~$\epsilon$:
\begin{equation}
\label{eq:loss}
    \mathcal{L} = \mathbb{E}_{(\mathcal{I}, \mathcal{I}^*, \mathcal{P}),\,t,\,\epsilon\sim\mathcal{N}(0,1)} \Bigl[ \| \epsilon -  \epsilon_\theta\left(z_t, t, \mathcal{I}, \mathcal{P}\right) \|_2^2 \Bigr],
\end{equation}
where the expectation is over the input dataset (input $\mathcal{I}$ and target image $\mathcal{I}^*$, input prompt $\mathcal{P}$) and diffusion timesteps~$t$. $\epsilon$ is the actual noise, and $\epsilon_{\theta}(z_t,t,\mathcal{I},\mathcal{P})$ is the predicted noise by the model with parameters $\theta$ for input image $\mathcal{I}$ with input prompt $\mathcal{P}$ at timestep $t$ with noised latent $z_t$.

\paragraphcustom{Model inference and sampling initialization.}
To generate images, we use DDIM sampling \cite{song2020denoising} with classifier-free guidance applied to the text prompt \cite{ho2021classifier}. 
In contrast to the standard DDIM sampler, we initialize the sampling process from a noise-perturbed latent representation $z_t$ of the input image $\mathcal{I}$ instead of Gaussian noise $\mathcal{N}(0,1)$ \cite{meng2021sdedit}. This improves the quality of the results and makes the output image more consistent with the input scene, which is important in our case as we wish to preserve the background and the identity of the objects in the scene, yet transform their states. This is especially true for out-of-distribution scenes when the sampled noise represents an image far from the input image in the latent space. We demonstrate benefits of this initialization in Section~\ref{sec:results}.

\begin{figure*}[t]
\centering
\includegraphics[width=\textwidth]{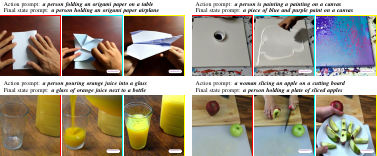}
  \vspace*{-6mm}
    \caption{\textbf{Various model predictions for moments from instructional videos unseen during training.} We generate the action (\textcolor{cyan}{blue}) and the final state of the object (\textcolor{yellow}{yellow}) given the initial state image (\textcolor{red}{red}) and the corresponding text prompt ({\bfseries{\textit{bold}}}) as the input. Our method correctly models hands interacting with objects (top left) and preserves scene elements such as the cutting board (bottom right). The method can also introduce tools, such as a knife, into the scene to fulfill the prompts, \eg, the slicing action (bottom right).}
    \vspace*{-2mm}
    \label{fig:additional_results}
\end{figure*}

\section{Experiments}
\label{sec:experiments}

We introduce our dataset and evaluation protocol in Sec.~\ref{sec:ds_experiments}. 
Then, in Sec.~\ref{sec:results}, we show our qualitative results and ablations.
Finally, we compare our model with related methods both qualitatively and quantitatively in Sec.~\ref{sec:comp_with_sota}.

\subsection{Dataset and experimental setup}\label{sec:ds_experiments}
\paragraphcustomWOvspace{Dataset.} Our dataset consists of 5-tuples extracted from videos of the ChangeIt and the COIN datasets (Section~\ref{subsec:dataset_acq}) using the self-supervised model~\cite{soucek2022multi}. 
Specifically, from each video, we select one 5-tuple of images and corresponding text descriptions,
$(\mathcal{I}, \mathcal{I}^*_{ac}, \mathcal{P}_{ac}, \mathcal{I}^*_{st}, \mathcal{P}_{st})$.
We remove images predicted with low confidence. 
As a form of augmentation, we train the self-supervised model multiple times with different seeds to select multiple different tuples from each video. In total, from 35k videos of the COIN and the ChangeIt datasets, we selected approx. 200k tuples.

We select five interaction categories of the ChangeIt dataset as held-out categories and use all other examples as the training set $\mathcal{D}_{train}$. From those five held-out categories, we manually select and verify 233 tuples as the test set $\mathcal{D}_{test}$. 
Additionally, we use all other tuples from the held-out categories as the held-out set $\mathcal{D}_{heldout}$, containing 14k tuples, for quantitative evaluation detailed in the next paragraphs. We make sure that tuples from the same video are not both in $\mathcal{D}_{test}$ and $\mathcal{D}_{heldout}$.

\paragraphcustom{Evaluation details.}
For quantitative evaluation, we report standard FID score~\cite{heusel2017gans}. However, we acknowledge its limitations~\cite{stein2023exposing}, and introduce a novel classification-based evaluation. 
We report accuracy on the fixed test set of real images~$\mathcal{D}_{test}$ from five held-out categories of the ChangeIt dataset.
For each category, we consider the initial and final states as separate classes, and thus have 10 classes in total (5 $\times$ 2). As the initial $\mathcal{I}$ and the final state $\mathcal{I}^*_{st}$ test images come from the same video, they contain the same background scene but only differ in the object state.
We train the classifier using real images for initial states ($\mathcal{I}$ from $\mathcal{D}_{heldout}$) but \textit{generated images for final states} (the output of a generative method conditioned on the initial state images and prompts $\mathcal{I}$, $\mathcal{P}_{st}$ from $\mathcal{D}_{heldout}$). 
Our intuition is that if the generated images (used for training) indeed properly represent the semantics of the final states, then the fixed test set (of real images) will be correctly classified. We also run the same evaluation for actions with the generated action images $\mathcal{I}^*_{ac}$.
Additionally, we perform a user study (\ARXIVversion{Appendix~\ref{supmat:userstudy}}{see the appendix~\cite{soucek24genhowto}}) to further verify the semantics and consistency of the generated images.


\begin{figure*}
\centering
\includegraphics[width=\textwidth]{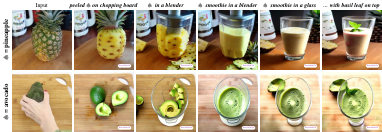}
  \vspace*{-6mm}
    \caption{\textbf{Long Term Generation.} Each image is generated recurrently using the image to the left and the prompt (top). The leftmost image is a real photo. Even though there are some compounding artifacts, especially when large changes to the scene are required (\eg, a blender $\to$ a glass), our model can generate plausible chains of transformations while preserving the scene.
    }
    \label{fig:statechain}
      \vspace*{-4mm}
\end{figure*}

\subsection{Results}\label{sec:results}

\paragraph{Qualitative results.} We show qualitative results of our method for various tasks in Figures \ref{fig:teaser}, \ref{fig:additional_results}, and \ref{fig:statechain}. The results show our model can make complex semantic changes to the images by (i)~faithfully modifying the objects according to the prompts, (ii)~correctly modeling hands interacting with the objects, (iii)~introducing new objects such as knives, and (iv)~preserving the background and other static parts of the scene from the input image. We show we not only correctly capture object transformations, but also actions, which are more difficult due to the involvement of other objects such as tools and human body parts.
While our method is trained on instructional videos only, we also show the model retains generalization ability stemming from the original Stable Diffusion weighs used for initialization. Figures in the \ARXIVversion{appendix}{appendix~\cite{soucek24genhowto}} show our method generating objects not available in our datasets, such as a toy, a parrot, or northern lights in a landscape photo.

\begin{figure}
\centering
\includegraphics[width=\linewidth]{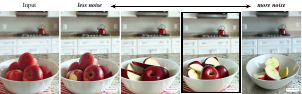}
\vspace*{-6mm}
    \caption{\textbf{Varying levels of noise used to initialize image generation process.} The rightmost image shows vanilla DDIM sampling initialized from Gaussian noise. Other images show when the sampling process is initialized from a noise-perturbed latent representation of the input image (left) with varying amounts of noise. Less noise generates images closer to the input scene. An initialization with the right amount of noise (black frame) better preserves the scene 
    while changing the state of the object. 
    }
    \label{fig:ddimabl}
    \vspace*{-3mm}
\end{figure}

\paragraphcustom{Ablations.} We ablate our decision to train two separate models for generating actions and for generating the final states. 
We show the quantitative results of training one joint or two separate models in Table~\ref{tab:ablation}. While we do not observe almost any drop in performance for the actions when using the joint model, the performance of the classifier for object states decreases significantly, coinciding with the drop in visual quality of the generated images.
We hypothesize that the additional visual complexity (\eg, correct placement of hands, \etc) necessary for correct action generation harms the visual fidelity of the generated object states.

We also provide a qualitative analysis of the DDIM sampling strategy. As discussed in Sec.~\ref{subsec:genhowto}, we initialize the sampling process from a noise-perturbed latent representation $z_t$ of the input image $\mathcal{I}$ instead of pure random noise $\mathcal{N}(0,1)$. In Figure~\ref{fig:ddimabl}, we show that starting purely from noise (the rightmost image) can lead to generated images that do not preserve well the original scene.
More examples as well as additional ablations on dataset acquisition are in \ARXIVversion{Appendix~\ref{supmat:dsexamples}}{the appendix~\cite{soucek24genhowto}}.

\begin{table}
  \centering
  {\small
  \begin{tabular}{lcc}
    \toprule
    Model training data  & Acc\textsubscript{ac} $\uparrow$ & Acc\textsubscript{st} $\uparrow$\\
    \midrule
    Only actions   & {0.77} & -      \\
    Only final states    & -      & {0.88} \\
    Joint model           & {0.76} & {0.79} \\
    \bottomrule
  \end{tabular}
  }
  \vspace*{-2mm}
  \caption{\textbf{Ablation of joint training.} Test set accuracy of a linear classifier trained on generated images and evaluated on a set of real images, measured for actions Acc\textsubscript{ac} and final states Acc\textsubscript{st} separately. Training separate models for actions and final states is better than a single joint model.}
  \label{tab:ablation}
    \vspace*{-2mm}
\end{table}

\subsection{Comparison with the state-of-the-art}\label{sec:comp_with_sota}

\paragraphcustomWOvspace{Compared methods.} We compare our method to various baselines as well as state-of-the-art methods for editing images based on text prompts: \textbf{(a) Stable Diffusion}~\cite{rombach2022high} is a text-conditioned model that only considers the text input without the input image. Our GenHowTo model is derived from stable diffusion hence we consider stable diffusion as an important baseline in our experiments.
\textbf{(b) Edit Friendly DDPM}~\cite{huberman2023edit} uses an off-the-shelf diffusion model (in our case, Stable Diffusion) but alters the sampling process to allow for more natural image editing.
\textbf{(c) InstructPix2Pix} \cite{brooks2023instructpix2pix} trains a conditional diffusion model that takes the input image $\mathcal{I}$ and the edit prompt $\mathcal{P}$ as an input to produce the modified image.
As the method is trained to follow instructions, we prepend the prompts with the phrase ``make it ...''.
We do not consider methods such as Imagic~\cite{kawar2023imagic} as they are not suited for large-scale image editing due to their immense time requirements just to edit a single image.

\begin{table}
  \centering
  {\small
  \begin{tabular}{c@{~~~}lcc}
    \toprule
    &Method & Acc\textsubscript{ac} $\uparrow$ & Acc\textsubscript{st} $\uparrow$ \\
    \midrule
    \multicolumn{4}{c}{\textit{test set categories unseen during training}}\\

    \textbf{(a)} & Stable Diffusion \cite{rombach2022high}                    & 0.51 & 0.50 \\
    \textbf{(b)} & Edit Friendly DDPM \cite{huberman2023edit}                 & 0.60 & 0.61 \\
    \textbf{(c)} & InstructPix2Pix \cite{brooks2023instructpix2pix}           & 0.55 & 0.63 \\
    \textbf{(d)} & \textit{CLIP (manual prompts)} \cite{radford2021learning}  & 0.52 & 0.62 \\
    \textbf{(e)} & \textbf{GenHowTo}                                          & \textbf{0.66} & \textbf{0.74} \\
    \midrule
    \multicolumn{4}{c}{\textit{test set categories seen during training}}\\
    \textbf{(f)} & Edit Friendly DDPM\textsuperscript{$\dagger$} \cite{huberman2023edit} & 0.69 & 0.80 \\
    \textbf{(g)} & \textbf{GenHowTo}\textsuperscript{$\dagger$}  & \textbf{0.77} & \textbf{0.88} \\
    \midrule
    \textbf{(h)} & \textit{Real images}                  & 0.96 & 0.97 \\
    \bottomrule
    \multicolumn{4}{c}{\textsuperscript{$\dagger$}~{\footnotesize Models trained also on the test set \textit{categories}.}}\\
  \end{tabular}
  }
\vspace*{-3mm}
  \caption{\textbf{Comparison with state-of-the-art using classification accuracy.} Test set accuracy of a linear classifier trained on generated images from various methods, measured for actions and final states separately.}
  \label{tab:cls}
  \vspace*{-3mm}
\end{table}

\paragraphcustom{Classification results.}
Table \ref{tab:cls} shows the comparison of different methods using the classification accuracy metric.
First, we compare the zero-shot capability of the methods by evaluating on categories unseen during training. As expected, a text-only diffusion model \textbf{(a)} performs poorly in the benchmark. It generates images visually distant from the input images as it does not have an image as input. Therefore, the learned classifier cannot distinguish correctly between the classes. The learned classifier does not differentiate between the object state classes (\eg, empty vs. full glass), instead, it differentiates between the real images and generated ones (all \textit{empty glass} training images are real, and all \textit{full glass} images are generated).  The same issue largely persists for Edit Friendly DDPM \textbf{(b)} or InstructPix2Pix \textbf{(c)}. We hypothesize the root cause is the aesthetic-oriented still images the methods have been trained with are very different from the scenes in instructional videos. In contrast, images generated by our method \textbf{(e)} better imitate the frames from real videos, yielding much-improved accuracy. As a baseline, we also measure the accuracy of the CLIP~\cite{radford2021learning} zero-shot classifier on the test set \textbf{(d)}---we conclude the CLIP baseline is not sufficient to distinguish and accurately classify small changes in the object appearance.

We also train our method as well as fine-tune Stable Diffusion on our dataset, including the held-out categories from $\mathcal{D}_{heldout}$ \textbf{(f--g)}. We observe a significant boost in performance when the models are tasked with generating types of objects and actions seen during training, significantly closing the gap with respect to the real images \textbf{(h)}, that serve as an upper bound.

\begin{table}
  \centering
  {\small
  \begin{tabular}{lcc}
    \toprule
    Method & FID\textsubscript{ac} $\downarrow$ & FID\textsubscript{st} $\downarrow$ \\
    \midrule
    \textit{Initial state images $\mathcal{I}$ from test set}  & 192 & 186 \\
    \midrule
    Stable Diffusion \cite{rombach2022high}           & 208 & 177 \\
    Edit Friendly DDPM \cite{huberman2023edit}        & 210 & 169 \\
    InstructPix2Pix \cite{brooks2023instructpix2pix}  & 223 & 173 \\
    \textbf{GenHowTo}                                 & \textbf{181} & \textbf{158} \\
    \bottomrule
  \end{tabular}
  }
  \vspace*{-2mm}
  \caption{\textbf{Comparison with state-of-the-art using the FID score.} Images generated using the initial state test set images $\mathcal{I}$ are contrasted with the real target action $\mathcal{I}^*_{ac}$ or final state $\mathcal{I}^*_{st}$ test set images. GenHowTo is the only method consistently generating images more similar to the target test set images than the input images (first row).}
  \label{tab:fid}
    \vspace*{-3mm}
\end{table}

\paragraphcustom{FID results.} The results in Table \ref{tab:fid} show that all methods generate final state images (column FID\textsubscript{st}) semantically more similar (lower FID score) to the real final state images than the real initial state images (Table \ref{tab:fid}, first row). However, in the case of actions (column FID\textsubscript{ac}), only our method has a lower FID score than the first row, \ie, it generates action images semantically more similar to the real action images than the real initial state images. 
This confirms the benefits of our approach that trains the generation model on real object transformations in instructional videos.

\begin{figure}[t]
\centering
\includegraphics[width=\linewidth]{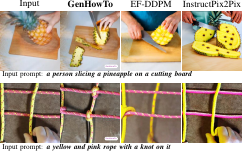}
    \vspace*{-7mm}
    \caption{\textbf{Qualitative comparison with related work.} We compare our method with Edit Friendly DDPM \cite{huberman2023edit} and InstructPix2Pix \cite{brooks2023instructpix2pix} on input images from videos unseen during training (first column) and edit prompts (below). The related methods struggle to preserve the background correctly. See the \ARXIVversion{appendix}{appendix~\cite{soucek24genhowto}} for additional examples.
    }
    \label{fig:relwork}
    \vspace*{-3mm}
\end{figure}

\paragraphcustom{Qualitative comparison with other methods.}
We qualitatively compare our method to other methods in Figure~\ref{fig:relwork}.
Both InstructPix2Pix and EF-DDPM often struggle to preserve background correctly. This can be seen especially in the first row, where both methods cannot preserve the chopping board from the input image. Also, the generated images of the related methods often look too artistic and unrealistic---we hypothesize it is due to the nature of the original ``artistic'' datasets those methods have been trained on. In contrast, our method both correctly preserves the background and realistically models the transformation of the object in the image.

\paragraphcustom{Limitations and societal impact.}
While our method can make complex semantic edits, it is based on trained models inheriting their limitations.
Image triplets selected for training (using~\cite{soucek2022multi}) can have weak spatiotemporal alignment. As a result, our method can struggle with some scenes and objects that move quickly, such as preserving the identity of people's faces or objects not represented well in the training data (see the \ARXIVversion{appendix}{appendix~\cite{soucek24genhowto}} for examples).
Additionally, our training preserves the qualities of the original Stable Diffusion weights, including its biases.
However, this allows for manipulations, such as painting style changes, when paired data is not available during our training.

\section{Conclusion}

In this work, we have introduced GenHowTo, an image- and text-conditioned generational model, that can generate images of actions and object state transformations while preserving the input image scene.
We trained GenHowTo on frames from instructional HowTo videos, automatically gathered by a recent self-supervised method, and prompts from an image captioning model.
We show our method is able to generate images that are more realistic with fewer hallucinations than recent competitors.
We believe this work can open up new avenues in generating intermediate goals for robotics and fine-grained action generation.

{\small
\paragraphcustom{Acknowledgements.}
This work was supported by the Ministry of Education, Youth and Sports of the Czech Republic through the e-INFRA CZ (ID:90140).
Part of this work was done within the University of Bristol's Machine Learning and Computer Vision (MaVi) Summer Research Program 2023. Research at the University of Bristol is supported by EPSRC UMPIRE (EP/T004991/1) and EPSRC PG Visual AI (EP/T028572/1). This research was co-funded by the European Union (ERC, FRONTIER, 101097822) and received the support of
the EXA4MIND project, funded by the European Union’s Horizon Europe Research and Innovation Programme, under Grant Agreement N° 101092944. Views and opinions expressed are however those of the author(s) only and do not necessarily reflect those of the European Union or the European Commission. Neither the European Union nor the granting authority can be held responsible for them.
}

{\small
\bibliographystyle{ieee_fullname}
\bibliography{egbib}
\flushcolsend
}

\ARXIVversion{\newpage\appendix\include{supmat.tex}}

\end{document}

%% file: supmat.tex
\ARXIVversion{\section*{Appendix}}{\section*{Overview}}

\ARXIVversion{The appendix}{This supplementary material} details the GenHowTo model training and dataset acquisition in Section \ref{supmat:implementation}.
Then, in Section \ref{supmat:dsexamples}, we provide additional insights into our method and dataset choices. In Section~\ref{supmat:userstudy}, we compare GenHowTo to previous methods through a user study. Finally, in Section \ref{supmat:qualitativeResults}, we show a large variety of qualitative results.

\section{Additional details}\label{supmat:implementation}
\paragraphcustomWOvspace{GenHowTo details.} We train the GenHowTo model for four days on eight A100 GPUs with a batch size of 32 at the resolution of 512$\times$512 pixels. We build on the official implementation of the ControlNet \cite{zhang2023adding}. The model is initialized with the Stable Diffusion v2.1 base EMA weights and trained using a fixed learning rate of $2\cdot10^{-5}$. During inference, we use DDIM sampler with 50 denoising steps. For better consistency of the qualitative results with the input images, we skip the first two steps and instead use the noise-perturbed representation of the input image (see Sec.~\ARXIVversion{\ref{subsec:genhowto}}{3.3}).

\paragraphcustom{Evaluation details.} We provide the exact details to replicate our classification-based evaluation, including the way to obtain the test images on the project's GitHub website\footnote{\url{https://github.com/soCzech/GenHowTo}}. For the FID evaluation, we use the publicly available PyTorch implementation\footnote{\url{https://github.com/mseitzer/pytorch-fid}}. For each input image $\mathcal{I}$ and test prompt $\mathcal{P}$ from the test set, we generate three images with different random seeds. We compute the FID between the generated images and the real target images $\mathcal{I}^*$ from the test set. We report the results separately for the action images and the final state images.

\begin{figure}
\centering
\includegraphics[width=\linewidth]{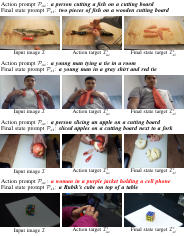}
    \caption{\textbf{Example of training dataset 5-tuples.} All shown images are automatically detected from instructional videos by \cite{soucek2022multi} and then automatically labeled using BLIP2. As the examples show, the images are nicely aligned across both the action and the final state, allowing our model to learn to manipulate images from instructional videos. However, as both the images and prompts are obtained automatically, they can contain errors (in red).}
    \label{supmat:training_data}
\end{figure}

\begin{figure*}
  \centering
  \begin{subfigure}{0.495\linewidth}
    \centering
\includegraphics[width=\linewidth]{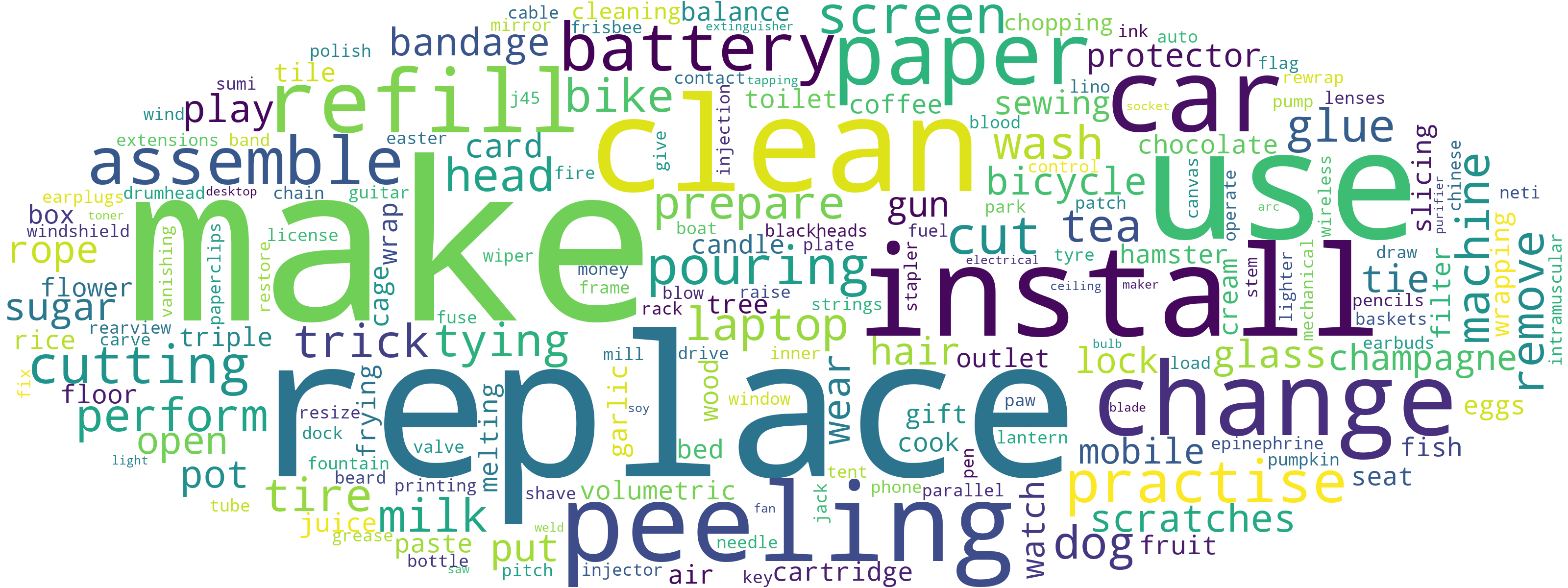}
    \vspace*{-3mm}
    \caption{\textbf{Wordcloud of interaction category names.}}
    \label{supmat:cloud_class}
  \end{subfigure}
  \hfill
  \begin{subfigure}{0.495\linewidth}
    \centering
\includegraphics[width=\linewidth]{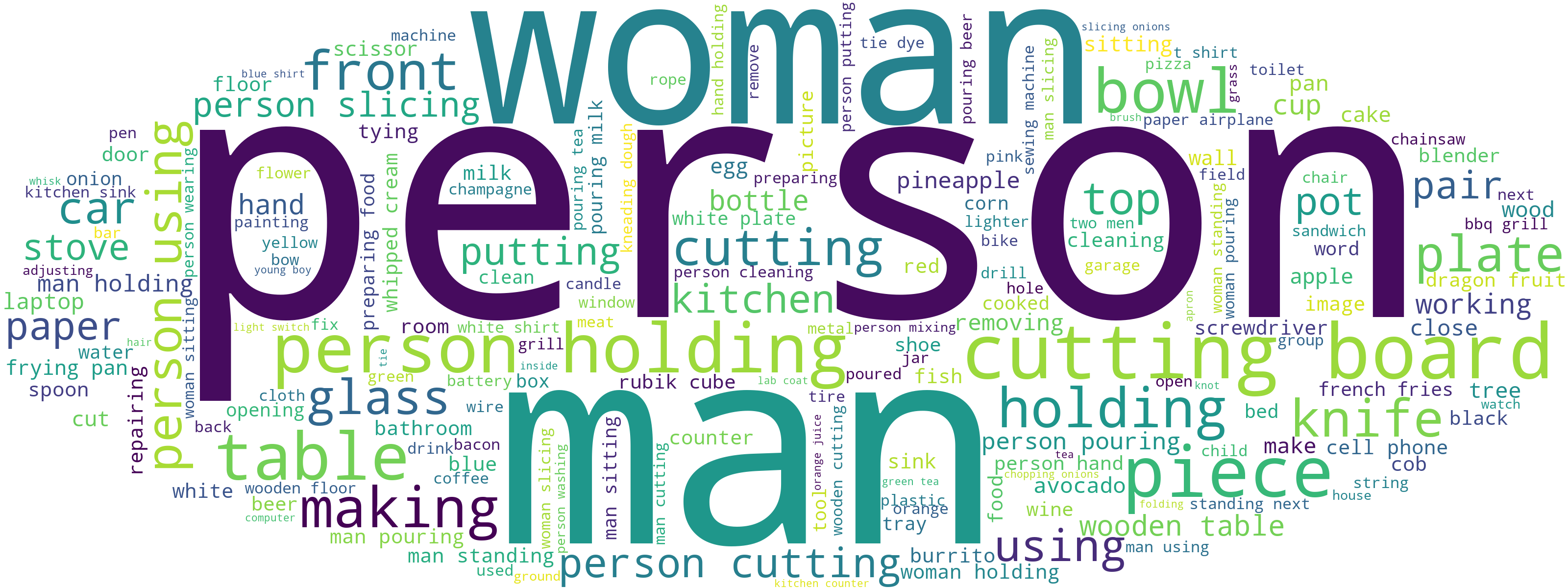}
    \vspace*{-3mm}
    \caption{\textbf{Wordcloud from automatically generated prompts.}}
    \label{supmat:cloud_blip}
  \end{subfigure}
\vspace*{-2mm}
  \caption{{The distribution of interaction categories from the ChangeIt and the COIN datasets used to train our model (left). And the~distribution of words from the automatically generated prompts for our target images extracted from the ChangeIt and the COIN datasets (right).}}
\vspace*{-3mm}
\end{figure*}

\paragraphcustom{Dataset aquisition details.} To obtain the initial state, action and final state images, we train the self-supervised model using the official implementation \cite{soucek2022multi}. We train separate models for the ChangeIt and for the COIN datasets. For each dataset, we train the model three times, starting with different random seeds. From each training, we select weights from three best-performing epochs---yielding nine model instances for each of the datasets. Each video from the two datasets is processed by the respective models, totaling nine correlated, yet often distinct, image sets, each containing the initial state, action, and final state per video. We keep approximately 90\% of COIN triplets with the highest classification scores as determined by the models to filter out incorrect predictions. For the ChangeIt dataset, we keep only 30\% as those videos are uncurated, and a large portion of the dataset consists of irrelevant videos.

For the text prompts, we use the zero-shot image-to-text version of BLIP2 weights \texttt{pretrain\_flant5xxl}. When generating the image captions, we leave the text prompt empty---we observed a reduction in the quality of the output caption when using text prompts. 

Sample 5-tuples of three images and two prompts can be seen in Figure \ref{supmat:training_data}. As the figure shows, the images are often well spatially aligned with only the object changing. This allows our model to learn to transform objects in the images according to prompts while keeping the background the same. In some cases, the automatically generated results can have errors (Figure \ref{supmat:training_data}, last row). In this case, the automatically generated caption for the action is incorrect.

\begin{figure}[t]
\centering
\includegraphics[width=\linewidth]{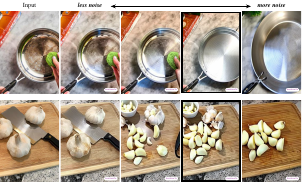}
    \vspace*{-5mm}\caption{\textbf{Varying levels of noise used to initialize image generation process.} The rightmost image shows vanilla DDIM sampling initialized from Gaussian noise. Other images show when the sampling process is initialized from a noise-perturbed latent representation of the input image (left) with varying amounts of noise. Less noise generates images closer to the input scene. An initialization with the right amount of noise (black frame) better preserves the scene (e.g. viewpoint -- top or color of the wooden board -- bottom) while changing the state of the object. 
    }
    \label{supmatfig:ddimabl}
\end{figure}
\begin{figure}
\centering
\includegraphics[width=.88\linewidth]{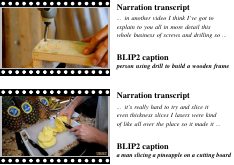}
    \caption{\textbf{Comparison of narration transcript vs. automatic captions.} We show approximately 8 seconds of ASR transcript centered around the shown video frame and the BLIP2 caption for the same video frame. We see the transcript often mentions information not shown in the video. The BLIP2 caption, on the other hand, describes what is happening in the frame.}
    \label{supmat:narrations_fig}
    \vspace*{-5mm}
\end{figure}
\begin{figure*}[thb!]
\centering
\includegraphics[width=\linewidth]{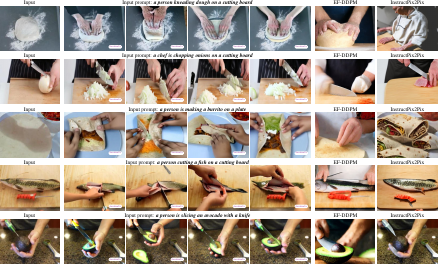}
    \caption{\textbf{Examples of GenHowTo action predictions.} We show multiple predictions per input, each with a different random seed. For comparison, we also show predictions of the related methods EF-DDPM and InstructPix2Pix in the last two columns. Our model correctly generates the person's hands interacting with the object in the scene. Also, our method can correctly preserve the background, which is not always true for the related methods.}
    \vspace{-2mm}
    \label{supmatfig:actionpred}
\end{figure*}
\begin{figure*}[thb!]
\centering
\includegraphics[width=\linewidth]{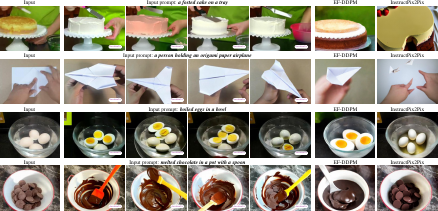}
    \caption{\textbf{Examples of GenHowTo final state predictions.} We show multiple predictions per input, each with a different random seed. For comparison, we also show predictions of the related methods EF-DDPM and InstructPix2Pix in the last two columns.}
    \label{supmatfig:statepred}
\end{figure*}

\paragraphcustom{Illustrating the variety of actions and objects.}
Our images come from instructional videos of the COIN and the ChangeIt datasets. Those datasets contain over two hundred interaction categories. The names of the interaction categories are shown as a word cloud in Figure \ref{supmat:cloud_class}. Similarly, we also show the distribution of words from the automatic image captions in Figure \ref{supmat:cloud_blip}. Together, the dataset images represent people manipulating objects in various environments. A large focus of the dataset is given to the most common environment in which people seek advice in---the kitchen.

\section{Additional ablations}\label{supmat:dsexamples}

\paragraphcustomWOvspace{Varying levels of noise.}
In Figure \ref{supmatfig:ddimabl}, we show more qualitative examples of our DDIM sampling strategy where we start with a noise-perturbed latent representation of the input image. In the example of the pan cleaning (top), our method correctly generates the transformed object (\ie, cleaned pan), yet in the case of random noise initialization (right), the method generates the object incorrectly scaled. In the second example of the peeled garlic, the image (right) generated only from random noise contains a darker background.
Here, even though our model sees the input image (via the ControlNet branch of the model), it may not be enough for perfect background reconstruction, as slightly misaligned training image pairs and other common video artifacts can force the model to focus more on the semantics and less on the pixel-level reconstruction. Starting from the noise-perturbed latent representation of the input image (Figure \ref{supmatfig:ddimabl}, black frame) corrects for these errors. Some misaligned training images are shown in Figure~\ref{supmat:training_data} (the Rubik's cube example).

\paragraphcustom{Automatically mined video frames.}
We investigate the significance of the unsupervised model~\cite{soucek2022multi} used in dataset acquisition. We test this by substituting our mined video frames with uniformly sampled frames. On our classification benchmark, the accuracy drops from $0.74$ (\ARXIVversion{Table~\ref{tab:cls}}{Table 2 in the main paper}, (e)) to $0.67$. We also observe poor visual quality and consistency of the generated images. This showcases the importance of our unsupervised mining approach for successfully training GenHowTo.

\paragraphcustom{Comparison of narration transcript vs. captions.} In our method, as described in Section \ARXIVversion{\ref{subsec:dataset_acq}}{3.2 in the main paper}, we refrain from using narration transcripts. Instead, we utilize an image captioning model to provide captions for our method. In Figure~\ref{supmat:narrations_fig}, we show a comparison between the narration transcript and the BLIP2 captions used in our work. We can observe the phenomenon already reported by \cite{Han_2022_CVPR,Miech_2019_ICCV} that the automatic narration is often noisy and may not align well with the content shown in the video.
Further, we also verify the claim quantitatively. We replace our BLIP-2 generated captions with the automatically obtained video narration (ASR) closest to the target frame. The accuracy on our classification benchmark drops from $0.74$ (\ARXIVversion{Table~\ref{tab:cls}}{Table 2 in the main paper}, (e)) to $0.60$.

\section{User study}\label{supmat:userstudy}

We run a user study by recruiting 10 people to assess generated images from GenHowTo, InstructPix2Pix~\cite{brooks2023instructpix2pix}, or EF-DDPM~\cite{huberman2023edit}.
In each case, we use the same initial image, and generated images from one of GenHowTo and one of the other two baselines (InstructPix2Pix or EF-DDPM).
Each rater was then shown the real initial state image, a question, and two generated images, in a random order. 
Raters had a forced choice of which method better addresses the questions.
\textcolor{red}{Q1}: \textit{``Which image better represents the final state described as \texttt{<input\_prompt>} of the same object as in the first image?''}. This question verifies the generation of the correct final state for the foreground object.
\textcolor{red}{Q2}: \textit{``Which image better preserves the consistency of the scene?''} to verify how well the methods preserve the background.

\begin{figure}[t]
\centering
\includegraphics[width=\linewidth]{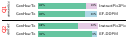}
    \vspace*{-6mm}
    \caption{\textbf{User preference between various methods.} Users prefer GenHowTo's outputs for being more true to the input prompt semantically as well as for keeping the background consistent with the original input image.
    }
    \label{fig:userstudy}
    \vspace*{-2mm}
\end{figure}

Figure \ref{fig:userstudy} shows how often each method was preferred, averaged over all 10 raters. GenHowTo consistently outperforms the other two methods for both questions. This showcases GenHowTo can better represent the semantics of the final state and better reserve the context.

\section{Additional qualitative results}\label{supmat:qualitativeResults}

\begin{figure}[t]
\centering
\includegraphics[width=\linewidth]{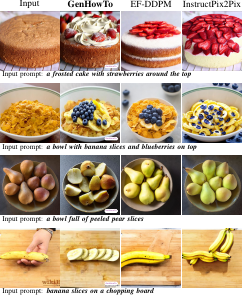}
    \caption{\textbf{Additional qualitative comparison with related work.} We compare our method with Edit Friendly DDPM \cite{huberman2023edit} and InstructPix2Pix \cite{brooks2023instructpix2pix} on out-of-distribution input images from the internet (first column) and edit prompts (below each row of images). Our dataset does not contain strawberries, blueberries, pears, or bananas, yet our method can correctly transform the objects according to the prompts. In contrast, the related methods often fail to transform the object. EF-DDPM also struggles to preserve the background of the input image.
    }
    \vspace{-3mm}
    \label{supmatfig:relwork}
\end{figure}

\begin{figure*}[!htb]
\centering
    \includegraphics[width=\linewidth]{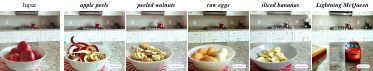}
    \caption{\textbf{A bowl of apples transformed in various ways.} Note that objects such as apples or eggs are part of the training data; however, bananas or child toys are missing. The model's ability to generate such objects comes from the initialization of our model with the StableDiffusion weights. To generate these images, we used the full prompt \textit{``[A bowl full of] ... on a kitchen countertop.''}}
    \label{supmatfig:var1}
\end{figure*}

\begin{figure*}[!htb]
\centering
    \includegraphics[width=\linewidth]{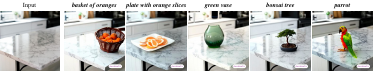}
    \caption{\textbf{Various objects added onto a marble countertop.} The model can generalize to objects not seen during our training using frames from instructional videos (\eg, vases, trees, parrots, \etc). The images were generated with the prompt \textit{``... on a marble countertop.''}} 
    \label{supmatfig:var2}
\end{figure*}

\begin{figure*}
\centering
    \includegraphics[width=\linewidth]{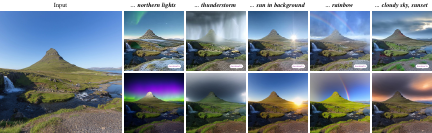}
    \caption{\textbf{Comparison of our method \textnormal{(top)} and InstructPix2Pix \textnormal{(bottom)} on a photo of landscape.} While InstructPix2Pix focuses more on changing the style of the photo, our method only applies more localized edits -- this is the result of the different training datasets.
    Note that our method has not been fine-tuned with any landscape photos, yet it is able to generalize, possibly due to the initialization from the StableDiffusion weights.}
    \label{supmatfig:iceland}
\end{figure*}

\paragraphcustomWOvspace{Additional qualitative results.}
We show additional qualitative results in Figures \ref{supmatfig:actionpred}, \ref{supmatfig:statepred}, \ref{supmatfig:relwork}, \ref{supmatfig:var1}, \ref{supmatfig:var2}, and \ref{supmatfig:iceland}. We show our method can correctly transform objects according to the input prompts. In Figures \ref{supmatfig:actionpred} and \ref{supmatfig:statepred}, we also show that our method generates a large variety of possible output images consistent with the input using different random seeds.

\paragraphcustom{Generalization ability.}
Before training, our method is initialized by pretrained Stable Diffusion weights. This gives our method some generalization ability, as shown in Figures~\ref{supmatfig:var1}, \ref{supmatfig:var2}, and \ref{supmatfig:iceland}. To the best of our knowledge, neither ChangeIt nor COIN datasets contain children's toys, parrots, or landscape scenes, yet our method can still produce meaningful results for such scenes and/or prompts.

\paragraphcustom{Additional qualitative comparison with related work.}
We show additional comparison with related work on input images from instructional videos in Figures \ref{supmatfig:actionpred}, \ref{supmatfig:statepred}, and input images from the internet in Figure \ref{supmatfig:relwork}. We can see our method not only works well on in-distribution data (Figures \ref{supmatfig:actionpred}, \ref{supmatfig:statepred}) but also outperforms the related methods on out-of-distribution objects such as strawberries or bananas---the objects not represented in the ChangeIt and COIN datasets.
Our method can also produce more localized edits than the closely related InstructPix2Pix method (Figure \ref{supmatfig:iceland}). The images generated by InstructPix2Pix focus more on the style of the output photo rather than only manipulating the target object(s).

\begin{figure*}
\centering
\includegraphics[width=\linewidth]{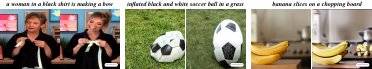}
    \caption{\textbf{Failure cases.} The performance degrades for always moving objects, such as faces (here not preserving the identity, left). The model can also struggle to preserve fine textures like grass (middle). Lastly, visual quality can suffer for objects not in the training data, especially in extreme viewpoints (right). Each example shows the input image (left), the input prompt (top), and the model's output (right).}
    \label{fig:failures}
\end{figure*}

\paragraphcustom{Limitations and failure modes.}
We show the limitations of our method, as described in \ARXIVversion{Section \ref{sec:comp_with_sota}}{the main paper}, in Figure \ref{fig:failures}.
Our method can struggle, \eg, with people's faces as they always move in the videos; therefore, they are not spatially aligned in our sets of input images containing the initial state, action, and final state.
Also, our model can sometimes ignore fine-grained textures. We hypothesize that the model focuses more on image semantics because some of the training images are not perfectly aligned---forcing the model to ignore pixel-level details in the input image(s).
Finally, as the method is trained on images extracted from a limited set of videos, applying the method to novel objects or to unusual views may result in a degraded performance. 


